\documentclass[twocolumn]{svjour3}
\smartqed  
\usepackage{amsmath}
\usepackage{tikz}
\usepackage{bm}
\usepackage{bbm}
\usepackage{amssymb}
\usepackage{booktabs}
\usepackage{tabularx}
\usepackage{array}
\newcommand{\vet}[1]{\mathbf{#1}}
\newcolumntype{C}[1]{>{\centering\arraybackslash\hspace{0pt}}p{#1}}
\usepackage{xcolor}
\usepackage{multirow}
\usepackage{hyperref}
\hypersetup{
	colorlinks,
	citecolor=blue,
	filecolor=black,
	linkcolor=red,
	urlcolor=red
}

\usepackage{algorithm}
\usepackage[noend]{algpseudocode}
\usepackage{natbib}

\newcommand{\name}{MATE}%
\newcommand{\nameFull}{Multi-tAsk mulTi-labEl}%

\newcommand{\cut}[1]{}

\newcommand{\revision}[1]{\color{black}{#1}}
\newcommand{\revisionTwo}[1]{\color{black}{#1}}

\begin{document}

\title{Intra-Camera Supervised Person Re-Identification}

\author{Xiangping Zhu \and Xiatian Zhu \and 
	Minxian Li \and Pietro Morerio \and Vittorio Murino \and Shaogang Gong}


\institute{Xiangping Zhu \at
	Computer Vision Institute, School of Computer Science and Software Engineering,
	Shenzhen University, China.\\
	Pattern Analysis and Computer Vision (PAVIS), Istituto Italiano di Tecnologia, Genova, Italy.\\
	\email{xiangping.zhu2010@gmail.com}
	\and
	Xiatian Zhu \at
	Vision Semantics Limited, London, UK.\\
	\email{eddy.zhuxt@gmail.com}
	\and
	Pietro Morerio \at
	Pattern Analysis and Computer Vision (PAVIS), Istituto Italiano di Tecnologia, Genova, Italy. \\
	\email{pietro.morerio@iit.it}
	\and
	Minxian Li and Shaogang Gong \at 
	Queen Mary University of London, UK. \\
	\email{m.li@qmul.ac.uk, s.gong@qmul.ac.uk}
	\and
	Vittorio Murino \at
	Ireland Research Center, Huawei Technologies Co. Ltd., Dublin, Ireland.\\
	Department of Computer Science, University of Verona, Verona, Italy. \\
	Pattern Analysis and Computer Vision (PAVIS), Istituto Italiano di Tecnologia, Genova, Italy. \\
	\email{vittorio.murino@univr.it}
}

 \date{Received: date / Accepted: date}

\maketitle

\begin{abstract}
Existing person re-identification (re-id) methods mostly exploit a large set of 
cross-camera identity labelled training data.
This requires a tedious data collection and annotation process,
leading to poor scalability in practical re-id applications.
On the other hand unsupervised re-id methods do not need identity label information,
but they usually suffer from much inferior and insufficient model performance.
To overcome these fundamental limitations, 
we propose a novel person re-identification paradigm
based on an idea of {\em independent} per-camera identity annotation.
This eliminates the most time-consuming and tedious inter-camera identity labelling process,
significantly reducing the amount of human annotation efforts.
Consequently, it gives rise to a more scalable and more feasible
setting, which we call {\em Intra-Camera Supervised (ICS)} person re-id,
for which we formulate a \nameFull{} (\name{}) deep learning method.
%
Specifically, \name{} is designed for self-discovering the cross-camera identity correspondence
in a per-camera multi-task inference framework.
Extensive experiments demonstrate the cost-effectiveness superiority of
our method over the alternative approaches on three large person re-id datasets.
For example, \name{} yields 88.7\% rank-1 score on Market-1501
in the proposed ICS person re-id setting,
significantly outperforming unsupervised learning models
and closely approaching conventional fully supervised learning competitors.

\end{abstract}

\begin{figure} 
	\centering
	\includegraphics[width=.5\textwidth]{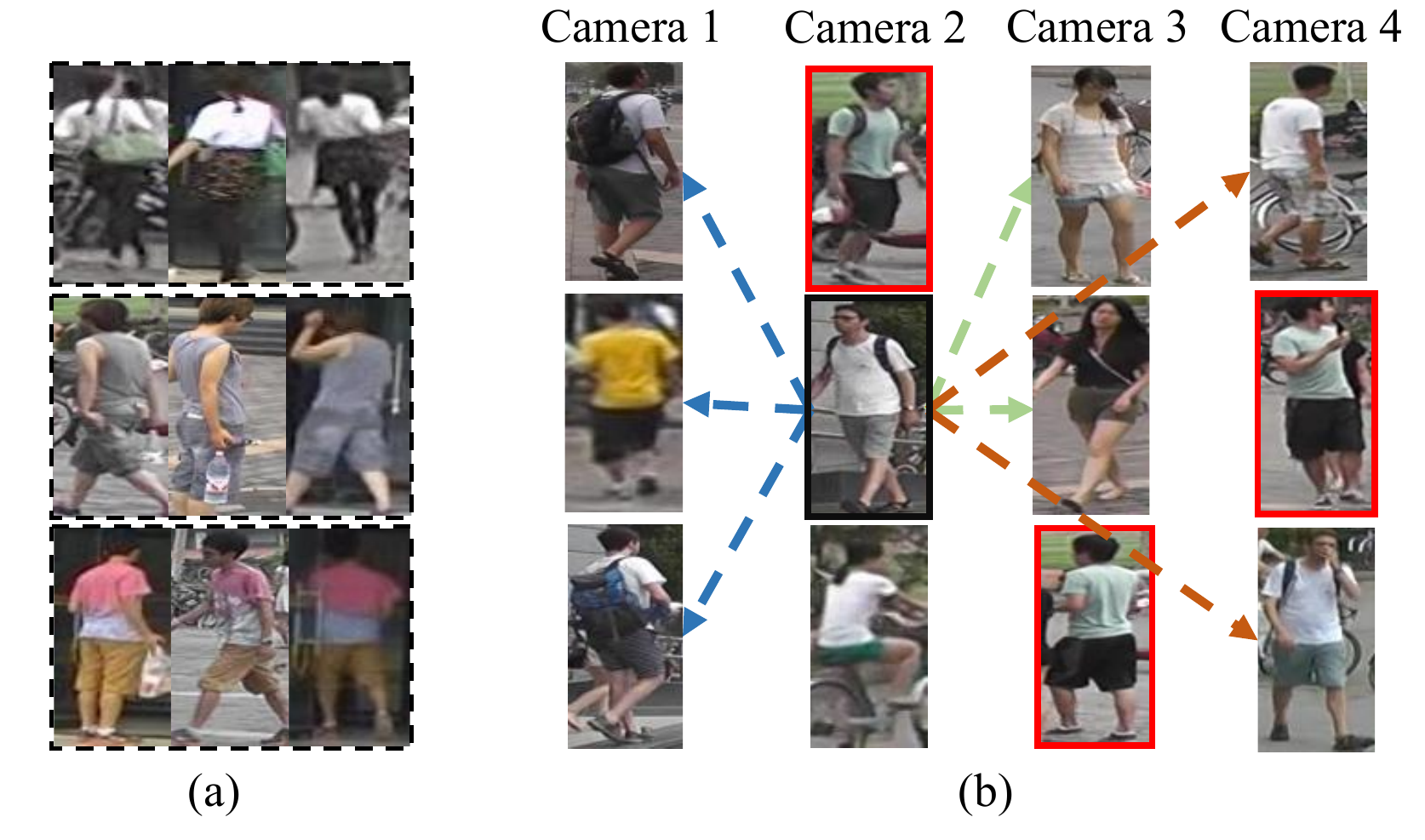}
	\caption{
		(\textbf{a}) Person re-identification challenges.
		Each triplet bounded by a dashed box shows the images of a single person
		from different camera views. 
		(\textbf{b}) Illustration of manually associating identities across camera-views. 
		The dashed arrow denotes the comparison between two identities.
		The associated identities are bounded with red boxes.	}
	\label{fig:domain_variations_inter_camera_annotation}
\end{figure}

\section{Introduction}
\label{sec:introduction}

Person re-identification (re-id) aims to
retrieve the target identity class
in detected person bounding box images 
captured by non-overlapping camera views
\citep{gong2014person,prosser2010svm_reid,sadf,dataset2014cuhk03,zheng2013reidentification}.
It is a challenging task due to the non-rigid structure of human body, 
highly unconstrained appearance variation across cameras, and the low resolution and low quality of the observations
(Fig. \ref{fig:domain_variations_inter_camera_annotation}(a)).
While deep learning methods 
\citep{chen2017person,li2018harmonious,sun2018beyond,Hou_2019_CVPR,zheng2019joint,zhou2019osnet}
have demonstrated remarkable performance advances,
they rely on supervised model learning from a large set of cross-camera 
identity labelled training samples.
This paradigm needs an exhaustive
and expensive training data annotation process
(Fig. \ref{fig:domain_variations_inter_camera_annotation}(b)),
dramatically lowering the usability while affecting the scalability of these methods 
for large scale deployment in real-world applications.

\begin{figure*}[t]
	\centering
	\includegraphics[width=0.9\textwidth]{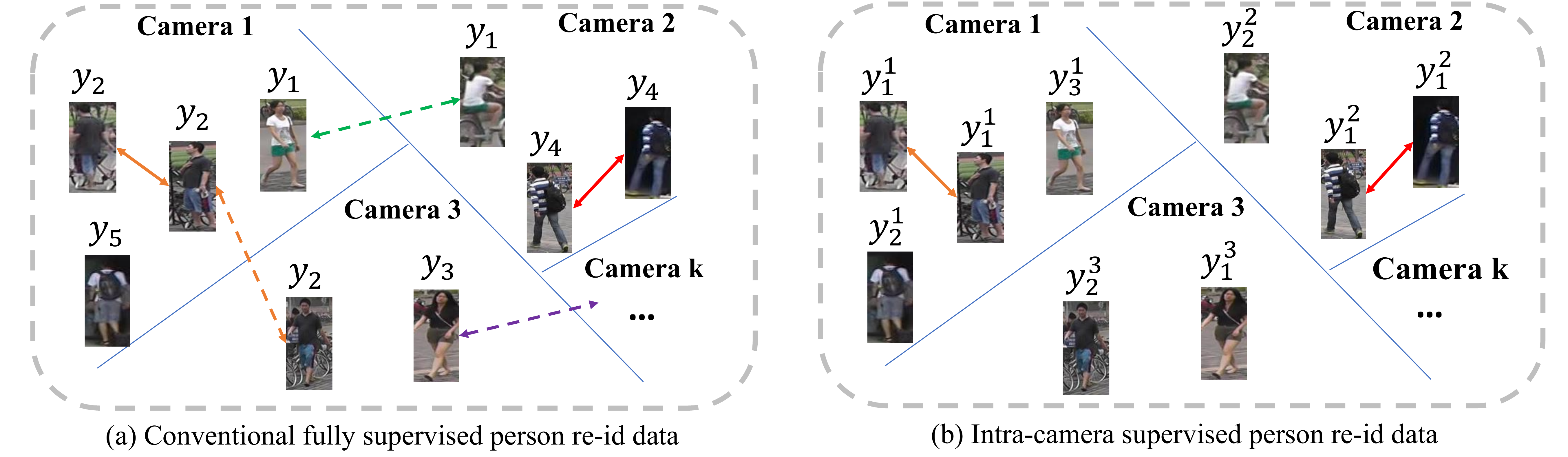}
	\caption{Labels in person re-id data.
		(\textbf{a}) Conventional fully supervised training data needs both {\em per-camera} and {\em cross-camera} 
		identity annotation in a unified class space.
		(\textbf{b})
		Intra-camera supervised (ICS) training data 
		only needs {\em per-camera} identity annotated {\em independently} in
		each camera view with a separate class space.
		Camera-view index is encoded as superscript of identity label in ICS person re-id data. 
		Solid and dashed arrows denote intra-camera and inter-camera association, respectively.
	}
	\label{fig:labelling}
\end{figure*}

\revision{
Specifically, for constructing a conventional person re-id training dataset, 
human annotators usually need to annotate person identity labels
both  
within individual camera views 
and across different camera views, 
and match a given person identity from one camera view
with all the persons from other camera views (inter-camera person identity association).
In particular, associating identity classes across camera views 
has a quadratic complexity with the number of both camera views and person identities
(Fig. \ref{fig:domain_variations_inter_camera_annotation}(a)).
This would significantly increase the cost of creating conventional training dataset. 

{\revisionTwo 
{ To quantify the annotation complexity}, we
{ consider} that (1) there are $N$ persons and $M$ camera views, 
	and (2) the cost of labelling every person is similar ({average cost}).
To label one person in {an} {\em intra-camera annotation}, {it requires} to compare this person with all the other unlabelled persons and the labelling complexity is $O(N)$.
So the labelling complexity for annotating all persons in one camera view is $O(N^2)$, and $O(MN^2)$ for all the camera views.	
For {an} {\em inter-camera annotation} (association), {we
  start with} the intra-camera labelling results. {Given} a
person identity from one camera-view,
	an annotator needs to compare it {exhaustively against the $N$
        identities from anyone of the other $M\!-\!1$ camera-views}, i.e., $N(M\!-\!1)$ identities.
This gives rise to a complexity of $O(N(M\!-\!1))$.
To label $N$ different persons, the {annotation complexity is} $O(N^2(M\!-\!1))$.
As not all persons would appear in every camera view in most cases,
this cross-camera view association needs to repeat for all $M$ camera
views, and the {actual cost can vary according to the proportion
  of people reappearing in pairs of camera views}.
{Therefore the inter-camera annotation complexity is between two
  extremes: $O(N^2M)$ for exhaustive reappearing}, and
  $O(M^2N^2)$ for {zero reappearing}. }
}

The problem of expensive training data collection has received significant attention.
Representative attempts 
for minimising the annotation cost
include: 
{\bf (1)} Domain generic feature design \citep{gray2008viewpoint,sadf,dataset2015market1501,xqda_lomo,gog}, 
{\bf (2)} Unsupervised domain adaptation \citep{uda_peng2016unsupervised,uda_deng2018image,uda_wang2018transferable,uda_lin2018multi,uda_zhong2018generalizing,uda_yu2018unsupervised,chen2019Instance},
{\bf (3)} Unsupervised image/tracklet model learning  \citep{wang2016towards,chen2018deep,lin2019BUC,li2019utal_pami,wu2020TSSL},
and
{\bf (4)} Weakly supervised learning \citep{ws_reid2019meng}.
By hand-crafting generic appearance features with prior knowledge, 
the {\em first} paradigm of methods can perform re-id matching universally.
However, their performances are often inferior 
due to limited knowledge encoded in such image representations.
This can be addressed by transferring the labelled training data
of a source dataset (domain), as demonstrated in the {\em second} paradigm 
of methods.
Implicitly, these methods assume
that the source and target domains share reasonably similar camera viewing conditions for ensuring sufficient transferable knowledge.
The heavy reliance on the relevance and quality of source datasets \citep{zhu2019UDA_attributes}
renders this approach less practically useful, since this assumption is often invalid. 
The {\em third} paradigm of methods is more scalable, 
as they need only unlabelled target domain data.
While having high potential, unsupervised re-id methods usually yield the weakest performance,
making them fail to meet the deployment requirements.
In contrast, the {\em fourth} paradigm of methods
considers a weakly supervised learning setting,
where the person identity labels are annotated at the video level
without fine-grained bounding boxes.
Apart from insufficient re-id accuracy, this paradigm is mostly sensible only when 
such weak labels can be cheaply obtained from certain domain knowledge,
which however is not generically accessible.

\begin{figure}[t]
\centering
\includegraphics[width=0.48\textwidth]{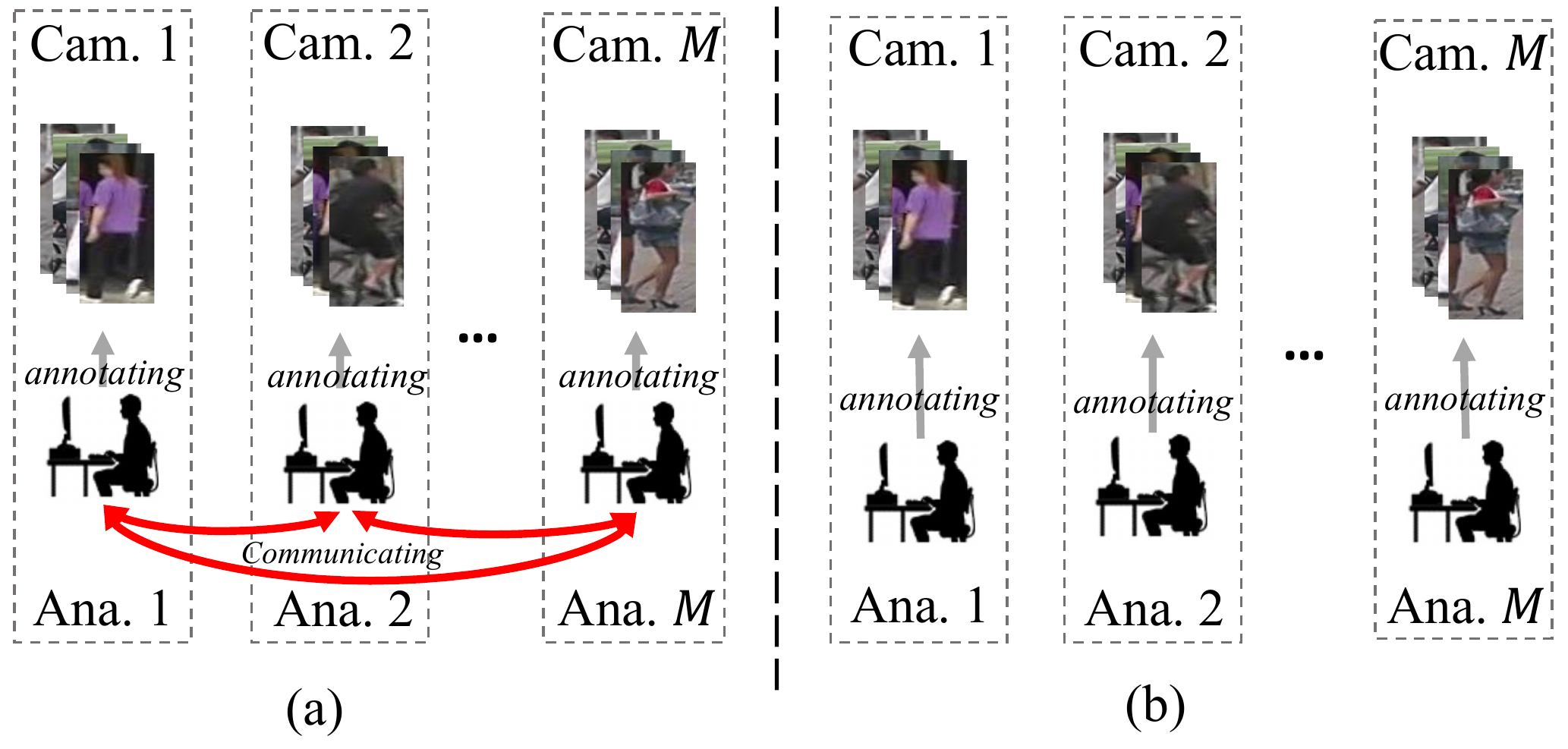}
\caption{Illustrations of data annotation process.
(\textbf{a}) Conventional fully supervised person re-id vs. (\textbf{b}) ICS person re-id
in the process of {\em training data collection}.
Suppose each annotator needs to label the training data from a different camera view.
In order to minimise the labelling conflict, 
an annotator may have to check if a person has been labelled or not
by others. 
This gives rise to expensive communication costs,
which is totally eliminated in the proposed ICS re-id paradigm,
due to the independence nature between camera views.  
}
\label{fig:annotating}
\end{figure}

In this work, we suggest another novel person re-identification
paradigm for scaling-up the model training process,
called \textbf{\em Intra-Camera Supervised} (ICS) person re-id
(Fig. \ref{fig:labelling}(b)).
As the name indicates, ICS eliminates the sub-process of cross-camera identity association
during annotation, 
which is the majority component of 
the standard annotation cost.
Under the ICS paradigm the training data involves only
the intra-camera annotated identity labels with each camera view 
labelled {\em independently}.
Importantly, as aforementioned, ICS naturally enables a parallel annotation process
by camera views without labelling conflict due to no cross-camera identity association
(Fig. \ref{fig:annotating}(b)).   
This desirable merit is lacking in the conventional training data labelling 
due to the difficulty of 
obtaining disjoint labelling tasks, e.g. subsets of person identity classes without overlap
(Fig. \ref{fig:annotating}(a)).
While being similar to the concurrent work \citep{ws_reid2019meng}
since they both consider explicitly the training data labelling process,
our ICS paradigm however does {\em not} assume specific domain knowledge therefore it is more generally 
applicable.

To solve the ICS re-id problem, we propose
a \textbf{\em \nameFull{}} (\name{}) deep learning model. 
Unlike the conventional fully supervised re-id methods using inter-camera identity labels,
\name{} is designed specially for overcoming two ICS challenges:
(1) how to learn effectively from per-camera independently labelled
training data, and 
(2) how to discover reliably the missing identity association across camera views.
Specifically, \name{} integrates two complementary learning components into a unified model:
{(a)} {\em Per-camera multi-task learning}
that separately learn individual
camera views for modelling their specificity and the implicit  
shared information in a multi-task learning manner
(Sec. \ref{subsec:multi_camera_multi_task_learning}).
This assigns a specific network branch (i.e. a learning task) for modelling each camera view
while constraining all the per-camera tasks to share a feature representation space.
{(b)} {\em Cross-camera multi-label learning}
that associates the identity labels
across camera views in a multi-label learning strategy
(Sec. \ref{subsec:cross_cam_learning}).
This is based on an idea of curriculum cyclic association
that can associate reliably multiple cross-camera identity classes from self-discovered
identity matches for multi-label model optimisation.

The {\bf contributions} of this work are:
{\bf (1)} We present a novel person re-identification paradigm 
for scaling up the model training process,
dubbed as {\em Intra-Camera Supervised} (ICS) person re-id. 
ICS is characterised by {\em no} need for exhaustive
cross-camera identity matching during training data annotation,
whilst allowing naturally parallel labelling
by camera views without conflict.
Consequently, it makes the training data collection substantially cheaper and faster
than the standard cross-camera identity labelling,
therefore offering a more scalable mechanism to large re-id deployments.
{\bf (2)} We formulate a {\em \nameFull{}} (\name{}) deep learning method for 
solving the proposed ICS person re-id problem.
In particular, \name{} combines the strengths of multi-task learning and multi-labelling learning 
in a unified framework to account for independent camera-specific 
identity label information and self-discovering their cross-camera association relationships
concurrently. This represents a natural strategy for fully leveraging the ICS supervision with
per-camera independent identity label spaces. 
{\bf (3)}
Through extensive benchmarking and comparisons
on the ICS variant of three large re-id datasets (Market-1501 \citep{dataset2015market1501}, 
DukeMTMC-reID \citep{dataset2017DukeMTMC-reID, dataset2016MTMC}, and MSMT17 \citep{dataset2018msmt17}), 
we demonstrate 
the cost-effectiveness advantages of the ICS re-id paradigm using our \name{} model
over the existing representative solutions including supervised learning,
semi-supervised learning, unsupervised learning, 
unsupervised domain adaptation, and tracklet learning.

A {\bf preliminary} version of this work was
published in \citep{zhu2019ics_reid}. 
Compared with this earlier study, there are a number of key differences:
{\bf (i)}
This study presents a more comprehensive investigation
into the proposed ICS person re-id paradigm in terms of training data annotation complexity,
along with a comparison to the standard cross-camera identity labelling method. 
This provides a more accurate measurement of training data collection cost,
revealing explicitly the intrinsic obstacles to scaling up model training
as suffered by the conventional supervised learning re-id paradigm.
{\bf (ii)}
We propose a more principled {\em \nameFull{}} learning method that can self-discover 
the cross-camera identity associations in a curriculum learning spirit. 
This improves dramatically the accuracy of cross-camera identity matching
and therefore the final model generalisation, 
as compared to the earlier method.
Besides, this new model performs unified end-to-end training
without the need for two-stage learning as required in the earlier version.
{\bf (iii)}
We provide more comprehensive evaluations and analyses of 
the ICS person re-id for giving holistic and useful insights,
in comparison to the existing alternative re-id paradigms.

\section{Related Work}
\label{sec:related_works}

{\bf Supervised person re-id }
Most existing person re-id models are created by {\em supervised} learning methods
on a separate set of cross-camera identity labelled training data
\citep{wang2014person,wang2016person,Zhao_2017_CVPR,chen2017person,li2017person,chen2018person,li2018harmonious,Song_2018_CVPR,chang2018MLFN_reid,sun2018beyond,Shen_2018_CVPR,dataset2018msmt17,Hou_2019_CVPR,zheng2019joint,Zhang_2019_CVPR,Wu_2019_CVPR,autoreid,zhou2019osnet}.
Relying on the strong supervision 
of cross-camera identity labelled training data,
they have achieved remarkable performance boost.
However, collecting such training data for each target domain
is highly expensive, limiting their usability and scalability 
in real-world deployments at scales.

\vspace{0.1cm}
\noindent{\bf Semi-supervised person re-id }
A typical strategy for supervision minimisation is by 
semi-supervised learning.
The key idea is to self-mine supervision information from unlabelled training data
based on the knowledge learned from a small proportion of labelled training data. 
A few attempts have been made in this research direction 
\citep{figueira2013semi,liu2014semi,wang2016towards,xin2019semi}.
However, this paradigm not only suffers from significant performance degradation
but also still needs a fairly large proportion of expensive cross-view pairwise labelling. 

\vspace{0.1cm}
\noindent{\bf Weakly supervised person re-id }
Recently, \citet{ws_reid2019meng} propose a weakly supervised 
person re-id paradigm where the identity labels are annotated
at the untrimmed video level. This setting makes more sense
when such identity labels are readily available from
certain domain knowledge which may be not generally provided. 
This is because, the major annotation cost of re-id training data
comes from matching identity classes across camera views,
rather than drawing person bounding boxes.
Often, person images are directly detected from the raw videos
by an on-the-shelf person detection model.
Therefore, this paradigm is not sufficiently general.

\vspace{0.1cm}
\noindent{\bf Unsupervised person re-id }
Unsupervised model learning is an intuitive solution to 
avoid the need of exhaustively collecting a large number
of labelled training data for every application domain.
Early hand-crafted feature based unsupervised learning methods
\citep{wang2014unsupervised,kodirov2015dictionary,kodirov2016person,khan2016unsupervised,ma2017person,
	ye2017dynamic,liu2017stepwise}
offer significantly inferior re-id matching performance,
when compared to the supervised learning counterparts.
Deep learning based methods \citep{lin2019BUC,wu2020TSSL} reduce this performance gap.
Besides, there are two research lines on unsupervised re-id learning
that become increasingly topical recently.

\vspace{0.1cm}
\noindent (1){\it Unsupervised domain adaptation } 

The key idea of domain adaptation based methods
\citep{uda_wang2018transferable,uda_fan18unsupervisedreid,peng2018joint,uda_yu2017cross,uda_zhu2017unpaired,deng2018image,uda_zhong2018generalizing}
is to explore the knowledge from the labelled data in
{\em related} source domains with model adaptation on the unlabelled target domain data.
Typical strategies include appearance style transfer 
\citep{uda_zhu2017unpaired,deng2018image,chen2019Instance},
semantic attribute knowledge transfer \citep{peng2018joint,uda_wang2018transferable},
and 
progressive source appearance information adaptation
\citep{uda_fan18unsupervisedreid,uda_yu2017cross}.
Although 
performing better than the
earlier unsupervised learning methods,
they require implicitly similar data distributions 
between the labelled source domain and the unlabelled target domain. 
This limits their scalability to arbitrarily
diverse (unknown) target domains in real-world deployments.

\vspace{0.1cm}
\noindent (2) {\it Unsupervised tracklet learning } 
Instead of assuming transferable source domain training data,
a small number of methods \citep{li2018taudl,li2019utal_pami,chen2018deep,wu2020TSSL} leverage the
auto-generated tracklet data with rich spatio-temporal information
for unsupervised re-id model learning. In many cases this is a feasible
solution as long  as video data are available. 
However, it remains highly challenging to achieve good model performance
due to noisy tracklets with unconstrained dynamics.

In this work, we introduce a new more scalable person re-id paradigm
characterised by intra-camera supervised (ICS) learning,
complementing the existing re-id scenarios as mentioned above.
In comparison, ICS provides a superior
trade-off between model accuracy and annotation cost,
i.e. higher cost-effectiveness.
This makes it a favourable choice
for large scale re-id applications with high accuracy performance requirement
and reasonably limited annotation budget.

\section{Problem Formulation}
\label{sec:prob_formulation}

We formulate the {\em Intra-Camera Supervised} (ICS) person re-identification problem. 
As illustrated in Fig. \ref{fig:labelling}(b),
ICS only needs to annotate intra-camera person identity labels {\em independently},
whilst eliminating the most-expensive inter-camera identity association
as required in the conventional fully supervised re-id setting.

Suppose there are $M$ camera views in a surveillance camera network. 
For each camera view $p \in \{ 1, 2, \cdots, M\}$, 
we {\em independently} annotate a set of training images
$\mathcal{D}^p = \{(\vet{x}_i^p, y_k^p)\}$
where each person image $\vet{x}_i^p$ is associated with an identity label $y_k^p\in \{y_1^p,y_2^p,\cdots,y_{N^p}^p\}$,
and $N^p$ is the total number of unique person identities in $\mathcal{D}^p$ \footnote{We use $i, j$ to denote image indexes, $k, l, t$ to denote identity indexes and $p, q$ to denote camera indexes.}.
For clarity, we express the camera view index in the superscript
due to the per-camera independent labelling nature in the ICS setting.
By combining all the camera-specific labelled data $\mathcal{D}^p$,
we obtain the entire training set as
$\mathcal{D} = \{\mathcal{D}^{1}, \mathcal{D}^{2}, \ldots, \mathcal{D}^{M}\}$.
%
%
For any two camera views $p$ and $q$, 
their $k$-th person identities $y_k^p$ and $y_k^q$ usually describe 
two different people,
i.e. they are two independent identity label spaces (Fig. \ref{fig:labelling}(b)).
This means exactly that the cross-camera identity association is {\em not} available,
in contrast to the fully supervised re-id data annotation (Fig. \ref{fig:labelling}(a)).

The ICS re-id problem presents a couple of new modelling challenges:
(1) how to effectively exploit the per-camera person identity labels,
%
and (2) how to automatically and reliably associate 
independent identity label spaces across camera views.
The existing fully supervised re-id methods do not apply 
due to the need for identity annotation in a {\em single} label space across camera views.
A new learning method tailored for the ICS setting is required to be developed.

\begin{figure*}[t]
\centering
\includegraphics[width=0.9999 \textwidth]{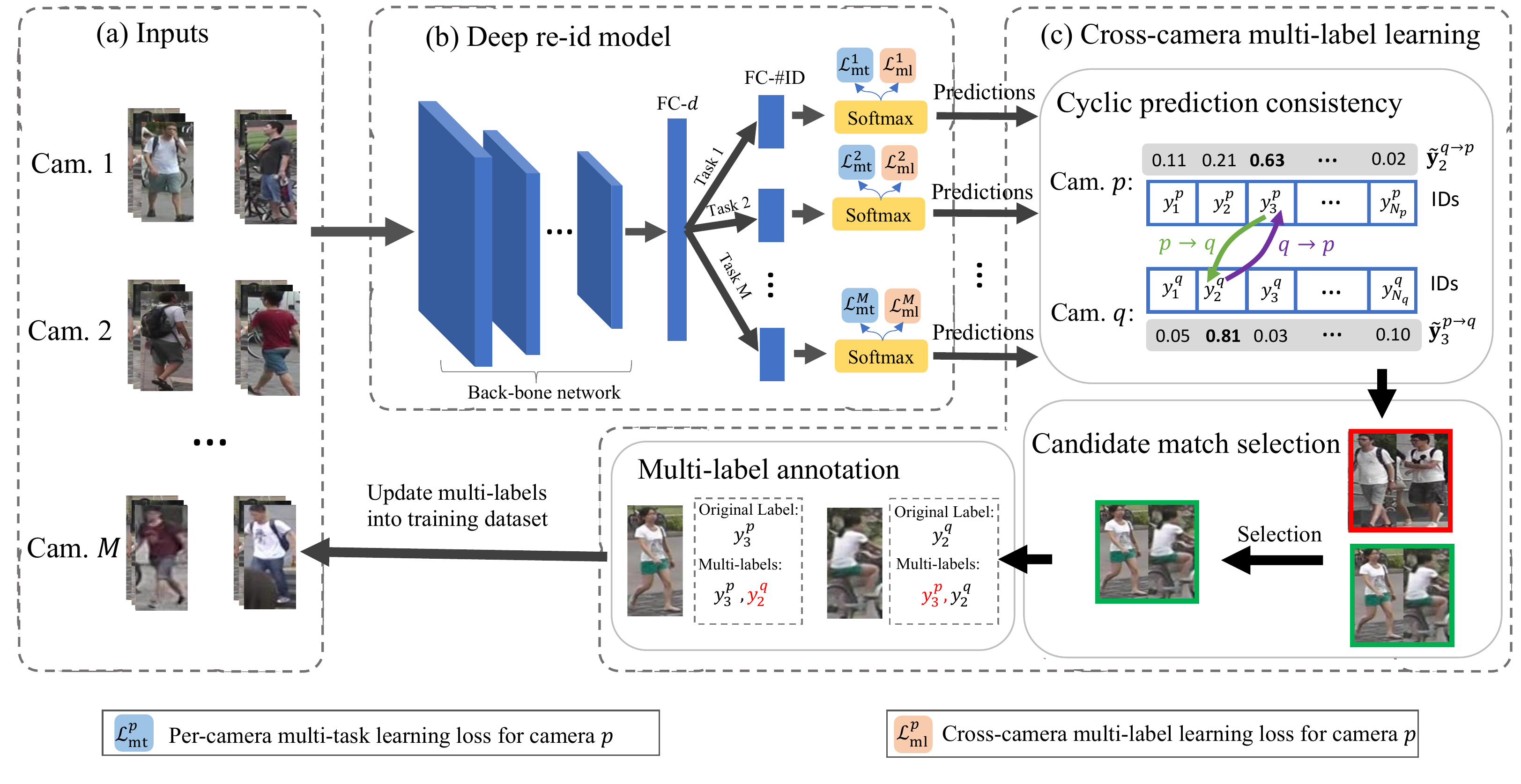}
\caption{Overview of the proposed \nameFull{} ({\bf \name{}}) deep learning method.
{\bf (a)} Given per-camera independently labelled training images,
\name{} aims to learn an identity discriminative feature representation model.
This is achieved by designing two learning components:
{\bf (b)} {\em Per-camera multi-task} learning
where we consider each individual camera view as a separate learning task with its own identity class space
and optimise these camera-specific tasks on a common feature representation
(Sec. \ref{subsec:multi_camera_multi_task_learning}),
and {\bf (c)} {\em Cross-camera multi-task} learning
where we self-discover the underlying identity matching relationships
across camera views via curriculum cyclic association and design a multi-label optimisation algorithm
to exploit these discovered cross-camera association information during model training.
The two components are integrated together in a single \name{} formulation,
resulting in an end-to-end trainable model.
}
\label{fig:pipeline}
\end{figure*}

\section{Method}
\label{sec:methodology}

We introduce a novel ICS deep learning method, 
capable of conducting {\bf M}ulti-t{\bf A}sk mul{\bf T}i-lab{\bf E}l ({\bf \name{}}) 
model learning 
to fully exploit
the independent per-camera person identity label spaces.
In particular, \name{} solves the aforementioned two challenges
by integrating two complementary learning components into a unified solution:
{\bf (i)} {\em Per-camera multi-task learning}
that assigns a separate learning task to each individual
camera view for dedicatedly modelling the respective identity space
(Sec. \ref{subsec:multi_camera_multi_task_learning}),
{\bf (ii)} {\em Cross-camera multi-label learning}
that associates the independent identity label spaces
across camera views in a multi-label strategy
(Sec. \ref{subsec:cross_cam_learning}).
Combining the two capabilities with a unified objective function,
\name{} explicitly optimises their mutual compatibility and complementary benefits
via end-to-end training.
An overview of \name{} is depicted in Fig. \ref{fig:pipeline}.

\subsection{Per-Camera Multi-Task Learning}
\label{subsec:multi_camera_multi_task_learning}

To maximise the use of {\em multiple} camera-specific identity label spaces
with some underlying correlation (e.g. partial identity overlap) in the ICS setting,
multi-task learning is a natural choice for model design \citep{argyriou2007multi}.
This allows to not only mine the common knowledge 
among all the camera views, but also to improve per-camera model learning concurrently
given augmented (aggregated) training data.

Specifically, 
given the nature of independent label spaces 
we consider each camera view as a separated learning task, 
all of which share a feature representation network
for extracting the common knowledge
in a multi-branch architecture design.
One branch is in charge of a specific camera view.
This forms {\em per-camera multi-task learning}
in the ICS context.
By such multi-task learning, our method can favourably derive a person re-id representation with {\em implicit} cross-camera identity discriminative capability, 
facilitating cross-camera identity association
\citep{li2019utal_pami}. This is because
during training, all the branches {\em concurrently} propagate the 
respective camera-specific identity label information
through the shared representation network $f_\theta$
(Fig. \ref{fig:pipeline}(b)),
leading to a {\em camera-generic} representation.
This process is done by minimising the softmax cross-entropy loss.

Formally, for a training image $(\vet{x}_i^p, y_k^p)\in \mathcal{D}^p$
from camera view $p$, 
the softmax cross-entropy loss is used for formulating the training loss:
\begin{equation}
\mathcal{L}_{\text{mt}}^p(i) =
{-} 
\mathbbm{1} (y_k^p)
{\log} \Big( g^p\big(f_\theta (\vet{x}_i^p)\big) \Big)
\label{eq:one_sample_mc_loss}
\end{equation}
where given the {\em camera-shared} feature vector 
$f_\theta (\vet{x}_i^p) \in \mathbbm{R}^{d\times 1}$,
the classifier $g^p(\cdot)$ for the camera view $p$ predicts
an identity class distribution 
in its own label space with $N_p$ classes:
$\mathbbm{R}^{d\times 1} \to \mathbbm{R}^{N_p\times 1}$.
The Dirac delta function $\mathbbm{1} (\cdot): \mathbbm{R} \to \mathbbm{R}^{1\times N_p}$ returns a one-hot vector with ``1'' at the specified index.

By aggregating the loss of training samples from all the camera views,
we formulate the {\em  per-camera multi-task learning} objective function
as:
\begin{equation}
\mathcal{L}_\text{mt}= \frac{1}{M} \sum_{p=1}^{M}
\Big( \frac{1}{B^p}\sum_{i=1}^{B^p} \mathcal{L}_{\text{mt}}^p(i)\Big)
\label{eq:mc_loss}
\end{equation}
where $B^p$ denotes the number of training images 
from the camera view $p$ in a mini-batch.

\subsection{Cross-Camera Multi-Label Learning}
\label{subsec:cross_cam_learning}

Cross-camera person appearance variation
is a key challenge for re-id.
Whilst this is implicitly modelled by the proposed multi-task learning as detailed above,
the per-camera multi-task learning is still insufficient 
to fully capture the underlying
identity correspondence relationships
across camera-specific label spaces.

However, it is non-trivial to associate identity classes across camera views.
One major reason is that a different set of persons may appear in a specific camera view,
leading to {\em no} one-to-one identity matching between camera views.
Conceptually, this gives rise to a very challenging open-set recognition problem where a rejection strategy is often additionally required \citep{open_set2013Scheirer1,open_set2013Scheirer2}.
Compared to generic object recognition in natural images, 
open-set modelling in re-id is more difficult 
due to small training data,
large intra-class variation, subtle inter-class difference,
and ambiguous visual observations of surveillance person imagery.
Besides, existing open-set methods
often assume accurately and completely labelled training data, and
the unseen classes only
in model test.
In contrast, we need to discover
cross-camera identity correspondences
during training with small (unknown) overlap across different spaces.

This is hence a harder learning scenario with a higher risk of error
propagation from noisy cross-camera association.
An intuitive solution for open-set recognition
is to find an operating threshold, e.g. by
Extreme Value Theory \citep{extreme_value_theory} based
statistical analysis.
This relies on optimal {\em supervised}
model learning from a sufficiently large 
training dataset, which however is unavailable in the ICS setting.

To circumvent the above problems, 
we design a \textbf{\em cross-camera multi-label learning} strategy
for robust cross-camera identity association.
This is realised by 
(i) designing a {\em curriculum cyclic association constraint}
to find reliable cross-camera identity association,
and (ii) forming a {\em multi-label learning algorithm}
to incorporate the self-discovered cross-camera identity association
into discriminative model learning
(Fig. \ref{fig:pipeline}({c})).

\subsubsection{Curriculum Cyclic Association}

For more reliable identity association across camera views,
we form a \textit{cyclic prediction consistency} constraint.
Specifically, given an identity class $y_k^p \in \{y_1^p, y_2^p,$ $\cdots,$ $y_{N^p}^p\}$ from a camera view
$p \in \{1, 2, \ldots, M\}$,
we need to find if a true matching identity (i.e. the same identity) exists in another camera view $q$.
We achieve this in the following process.

{\bf (i)} We first project all the images
of each person identity $y_k^p$ from camera view $p$ to 
the classifier branch of camera view $q$ to obtain a
{\em cross-camera prediction} $\tilde{\vet{y}}^{p \to q}_{k}$ via averaging as: 
\begin{align}
\tilde{\vet{y}}^{p \to q}_{k} = \frac{1}{S_{k}^p} \sum_{i=1}^{S_k^p} {g}^q\big(f(\vet{x}_i^p)\big)
\in \mathbb{R}^{N_q \times 1},
  \label{eq:cross_cam_map_prob}
\end{align}
where $S_{k}^p$ is the number of images from identity 
$y_k^p$. Each element
of $\tilde{\vet{y}}^{p \to q}_{k}$,
denoted as $\tilde{\vet{y}}^{p \to q}_{k}(l)$, 
means the identity class matching probability 
at which $y_k^p$ (an identity from camera view $p$) matches
$y_l^q$ (an identity from camera view $q$)
in a cross-camera sense.

{\bf (ii)}
We then nominate the person identity $y_{l^*}^q$ from camera view $q$
with the maximum likelihood probability as the candidate matching identity:
\begin{align}
l^{*} = \arg \max_{l} 
\tilde{\vet{y}}^{p \to q}_{i}(l), \; l \in \{ 1, 2, \ldots, N_q\}.
    \label{eq:cross_cam_match}
\end{align}
With such one-way ($p\to q$) association alone, 
the matching accuracy should be not satisfactory
since it cannot handle the cases of {\em no-true-match}
as typical in the ICS setting.  
To boost the matching robustness and correctness, 
we further design a 
\textit{curriculum cyclic association} constraint.

{\bf (iii)} Specifically, in an opposite direction of the above steps, 
we project all the images
of identity $y_{l^*}^q$ from camera view $q$ to 
the classifier branch of camera view $p$ 
in a similar way as Eq. \eqref{eq:cross_cam_map_prob},
and obtain the best candidate matching identity $y_{t^*}^p$ with Eq. \eqref{eq:cross_cam_match}.
Given this back-and-forth matching between camera view $p$ and $q$,
we subsequently filter the above candidate 
pair $(y_k^p, y_{l^*}^q)$ by a cyclic constraint as:
\begin{equation}
(y_k^p, y_{l^*}^q) 
\left\{ 
\begin{array}{ll}
\text{is a candidate match}, & \text{if} \;\; y_{t^*}^p = y_k^p, \\
\text{is not a candidate match}, & \text{otherwise}.
\end{array}
\right.   
\label{eq:match_pair}
\end{equation}
This removes non-cyclic association pairs.
While being more reliable, 
it is observed that only the cyclic association in
Eq. \eqref{eq:match_pair} is not sufficiently strong for {\em hard} 
cases (e.g. different people with very similar clothing appearance), 
leading to false association.

{\bf (iv)} To overcome this problem, inspired by the findings of cognitive study which suggest a better learning strategy is to start {\em small}
\citep{elman1993learning,krueger2009flexible},
we design a curriculum association constraint.
It is based on the cross-camera identity matching probability.
Formally, we define a {cyclic association degree} as:
\begin{align}
\psi^{p \Leftrightarrow q}_{k \Leftrightarrow l^*}
=
\tilde{\vet{y}}^{p \to q}_{k}(l^*) \cdot
\tilde{\vet{y}}^{q \to p}_{l^*}(k)
\label{eq:cad}
\end{align}
which measures the joint probability of a cyclic association
between two identities $y_k^p$ and $y_{l^*}^q$.
Given this unary measurement, we can deploy a {\em curriculum threshold}
$\tau \in [0, 1]$ for selecting candidate matching pairs via:
\begin{equation}
\text{Cyclic} \;\; (y_k^p, y_{l^*}^q)
\left\{
\begin{array}{ll}
\text{ is a match}, & \text{if } \psi^{p \Leftrightarrow q}_{k \Leftrightarrow l^*} > \tau, \\
\text{ is not a match}, & \text{otherwise}.
\end{array}
\right.
\label{eq:association_probability_constraint}
\end{equation} 
This filtering determines if a cyclically associated 
identity pair $(y_i^p, y_{k^*}^q)$ will be considered as a match.

\vspace{0.1cm}
\textbf{\em Curriculum threshold.}
The design of the curriculum threshold $\tau$ has a crucial influence 
on the quality of cross-camera identity association.
In the spirit of curriculum learning,
we consider $\tau$ as an annealing function
of the model training time to enable a progressive selection.
Meanwhile, we need to take into account that
the magnitude of maximum prediction usually increases 
along the training process as the model gets more mature. 
Taking these into consideration, 
we formulate the curriculum threshold as:
\begin{align}
\tau^r = \min\Big(\tau^u, \;\; \tau^l + \frac{r}{R-1} (1-\tau^l) \Big)
\label{eq:threshold}
\end{align}
where $r$ specifies the current training round,
with a total of $R$ rounds.
We maintain two thresholds: 
$\tau^u$, which denotes the upper bound, and 
$\tau^l$, which denotes the lower bound.
Both of these two thresholds can be estimated by cross-validation.

\vspace{0.1cm}
\textbf{\em Summary. }
We perform the above curriculum cyclic association process
for every camera view pairs, which outputs
a set of associated identity pairs across camera views.
This self-discovered pairwise information 
will be used to improve model training as detailed in the following.

\subsubsection{Multi-Label Learning.} 

To leverage the above identity association results
for improving model discriminative learning,
we introduce a multi-label learning scheme
in a cross-camera perspective. 
It consists of
(i) multi-label annotation
and 
(ii) multi-label training.

\vspace{0.1cm}
\textbf{\em (i) Multi-label annotation.}
For easing presentation and understanding, 
we assume two camera views,
and it is straightforward to extend to more camera views.
Given an associated identity pair $(y_k^p, y_{l^*}^q)$ obtained as above,
we annotate all the images $X_i^p$ of $y_i^p$ from camera view $p$ 
with an extra label $y_{l^*}^q$ of camera view $q$.
We do the same for all the images $X_{l^*}^q$ of $y_{l^*}^q$
in an inverse direction.
Both image sets are therefore annotated with the 
same two identity labels,
i.e. these images are associated.
See an illustration example in Fig. \ref{fig:pipeline}(c).
Given $M$ camera views, for each identity $y_k^p$ we perform
at most $M-1$ times such annotation 
whenever a cross-camera association is found,
resulting in a multi-label set $Y_i^p= \{y_k^p, y_{l^*}^q, \cdots\}$
for $X_i^p$,
with the cardinality $1 \leq |Y_i^p| \leq M$.
When $|Y_i^p|=1$, it means no cross-camera association is obtained.
When $|Y_i^p|=M$, it means an identity association is found
in every other camera view.

\vspace{0.1cm}
\textbf{\em (ii) Multi-label training. } 
Given such cross-camera multi-label annotation,
we then formulate a multi-label training objective 
for an image $\vet{x}_i^p$ as
\begin{equation}
\mathcal{L}_{\text{ml}}^p(i) =
\frac{1}{|Y_i^p|}
\sum_{y^c \in Y_i^p}
{-} \mathbbm{1} (y^c) {\log}\Big( g^c\big(f_\theta(\vet{x}_i^p)\big)\Big)
\label{eq:one_sample_ml_loss}
\end{equation}
where $c$ indices the camera view of $Y_i^p$
with the corresponding identity label
simplified as $y^c$.
For mini-batch training,
we design the cross-camera multi-label learning 
objective as:
\begin{equation}
\mathcal{L}_\text{ml}= 
\frac{1}{B} \sum_{i, p}\mathcal{L}_{\text{ml}}^p(i)
\label{eq:ml_loss}
\end{equation}
which averages the multi-label training loss
of all the $B$ number of training images
in a mini-batch.

\vspace{0.1cm}
\textbf{\em Remarks.}
It is noteworthy to point out that, in contrast to the conventional single-task multi-label learning \citep{tsoumakas2007multi},
we jointly form multi-label learning and multi-task learning in a unified framework,
with a unique objective of associating different label spaces and merging the
independently annotated labels with the same semantics. 

\subsection{Final Objective Loss Function}
By combining per-camera multi-task (Eq. \eqref{eq:mc_loss}) and cross-camera multi-label (Eq. \eqref{eq:ml_loss}) learning objectives, we obtain the final model loss function as:
\begin{align}
\mathcal{L} = \mathcal{L}_\text{mt} + \lambda \mathcal{L}_\text{ml},
\label{eq:mmm_loss}
\end{align}
where the weight parameter $\lambda\in [0, 1]$ is to trade-off the two loss terms. 
With this formula as model training supervision, 
our method can effectively learn discriminative re-id model 
using both camera-specific identity label spaces available under the ICS setting
($\mathcal{L}_\text{mt}$)
and cross-camera identity association self-discovered by \name{} itself
($\mathcal{L}_\text{ml}$)
concurrently.
The \name{} model training process is summarised in Algorithm \ref{algo:Algorithm}.

\begin{algorithm}[h]
	\caption{The \name{} model training procedure.} \label{algo:Algorithm}
	\textbf{Input:} Intra-camera independently labelled training data;\\
	\textbf{Output:} A trained person re-id model; \\
	\textbf{Model training:}\\ 
	\hphantom{~} 
	\textbf{for} $r=1$ \textbf{to}  \textsl{R} \textbf{do}: \\ 
	\hphantom{~~~}  
	Calculate the curriculum threshold $\tau^r$; \\
	\hphantom{~~~}  
	Cross-camera identity association as in Eqs. \eqref{eq:cross_cam_map_prob}-\eqref{eq:association_probability_constraint};\\
	\hphantom{~~~}
	\textbf{for} $e=1$ \textbf{to}  \textsl{epoch\_number} \textbf{do}: \\ 
	\hphantom{~~~~~~} 
	\textbf{for} $t=1$ \textbf{to}  \textsl{per-epoch mini-batch number} \textbf{do}: \\
	\hphantom{~~~~~~~~} 
	Feed forward a mini-batch of training images; \\
	\hphantom{~~~~~~~~}
	Compute learning loss using Eq. \eqref{eq:mmm_loss}; \\
	\hphantom{~~~~~~~~} 
	Update the network model by back-propagation; \\
	\hphantom{~~~~~~} 
	\textbf{end for} \\ 
	\hphantom{~~~}
	\textbf{end for} \\
	\hphantom{~}
	\textbf{end for} \\
\end{algorithm}

\section{Experiments}
\label{sec:exp}

{\bf Datasets. }
Due to {\em no} existing re-id datasets for the proposed scenario, 
we introduced three ICS re-id benchmarks.
We simulated the ICS identity annotation process
on three existing large person re-id datasets,
Market-1501 \citep{dataset2015market1501}, DukeMTMC-reID \citep{dataset2016MTMC,dataset2017DukeMTMC-reID} and MSMT17 \citep{dataset2018msmt17}. 
Specifically, 
for the training data of each dataset,
we {\em independently} perturbed the original identity labels
for every individual camera view, and ensured
that the same class labels of any pair of different camera views
correspond to two unique persons (i.e. no labelled cross-camera association).
We used the same original test data of each dataset
for model performance evaluation.

\vspace{0.1cm}
\noindent{\bf Performance metrics. }
Following the common person re-id works, the Cumulative
Matching Characteristic (CMC) and mean Average Precision (mAP) metrics were used for model performance measurement.

\vspace{0.1cm}
\noindent \textbf{Implementation details. }
The ImageNet pre-trained
ResNet-50 \citep{net_arc2016resnet} 
was selected as the backbone network of our \name{} model. 
As shown in Fig. \ref{fig:pipeline}, each branch in \name{} was formed 
by a fully connected (FC) classification layer. 
We set the dimension of the re-id feature representation to 512.
The person images were resized to $256\times 128$
in pixel. 
The standard stochastic gradient descent (SGD) optimiser was adopted.
The initial learning rate of the backbone network and 
classifiers were set to 0.005 and 0.05, respectively.
We set a total of 10 rounds to anneal the curriculum threshold $\tau$ (Eq. \eqref{eq:association_probability_constraint}),
with each round covering 20 epochs 
(except the last round where we trained 50 epochs to guarantee the convergence).
We empirically estimated 
$\tau^l=0.5$ (the lower bound of $\tau$)
and 
$\tau^u=0.95$ (the upper bound of $\tau$)
for Eq. \eqref{eq:threshold}.
In order to balance the model training across camera views, we randomly
selected from each camera the same number of images, i.e. 4 images, 
per identity and the same number of identities, i.e. 2 identities, 
to construct a mini-batch.
Unless stated otherwise, we set the loss weight $\lambda=0.5$ for Eq. \eqref{eq:mmm_loss}.
In test, the Euclidean distance was applied 
to the camera-generic feature representations
for re-id matching. 

\begin{table} [t]
 \center
 \caption{Benchmarking the ICS person re-id performance.
 }
 	\label{tab:benchmark}
	\begin{tabular}{c||cccc}
		\hline
		Dataset &\multicolumn{4}{c}{Market-1501} \\
		\hline
		Metric (\%)  & R1 & R10 & R20 & mAP \\
		\hline
		MCST
		& {34.9} & 60.1 & 69.3  & {16.7}\\
		EPCS
		& {42.6} & 64.6 & 71.2 & {19.6}\\
		PCMT
		& {78.4} & 93.1 & 95.7  & {52.1}\\
		\hline
		\bf \name{} (Ours)      
		& \bf{88.7} & \bf{97.1} & \bf{98.2} &\bf {71.1}\\
		\hline
		\hline
		Dataset &\multicolumn{4}{c}{DukeMTMC-reID} \\
		\hline
		Metric (\%)  & R1 & R10 & R20 & mAP \\
		\hline
		MCST
		& {25.0} & 50.1 & 58.8  & {16.3}\\
		EPCS
		& {38.8} & 58.9 & 64.6  & {22.1}\\
		PCMT
		& 65.2 & 81.1 & 85.6 & 44.7\\
		\hline
		\bf \name{} (Ours)       
		& \bf {76.9} & \bf{89.6} & \bf{92.3} &\bf {56.6}\\
		\hline
		\hline
		Dataset &\multicolumn{4}{c}{MSMT17} \\
		\hline
		Metric (\%)  & R1 & R10 & R20 & mAP \\
		\hline
		MCST
		& {12.1} & 26.3 & 33.0  & {4.8}\\
		EPCS
		& {16.8} & 31.5 & 37.4 & {5.4}\\
		PCMT
		& {39.6} & 59.6 & 65.7 & 15.9\\
		\hline
		\bf \name{} (Ours)        
		& \bf {46.0} & \bf{65.3} & \bf{71.1} &\bf {19.1}\\
		\hline
	\end{tabular}
\end{table}

\begin{figure*}[t]
	\centering
	\includegraphics[width=0.99 \textwidth]{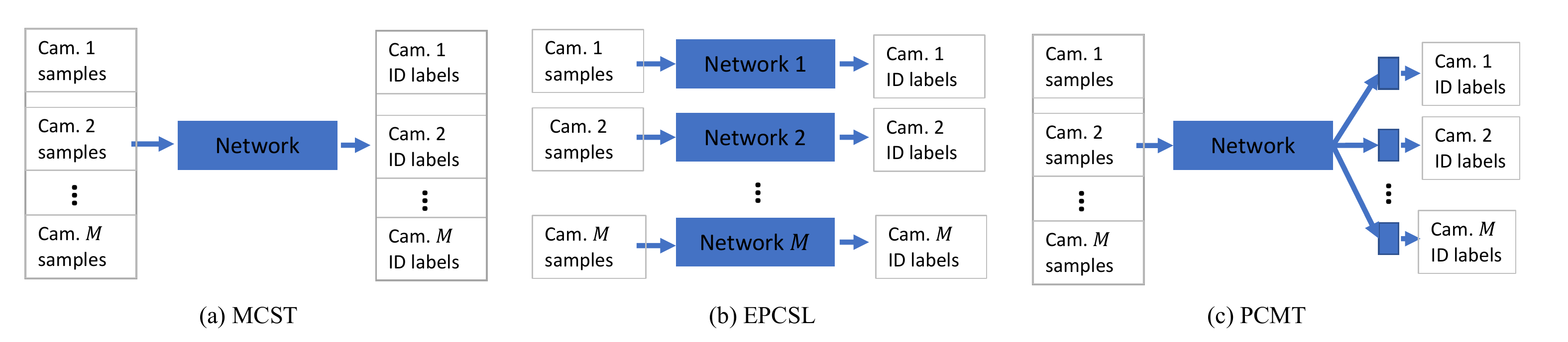}
	\caption{
		Three baseline learning methods for ICS person re-id: 
		(\textbf{a}) Multi-Camera Single-Task (MCST) learning. 
		(\textbf{b}) Ensemble of Per-Camera Supervised (EPCS) Learning. 
		(\textbf{c}) Per-Camera Multi-Task (PCMT) learning.
	}
	\label{fig:three_baselines}
\end{figure*}

\begin{figure*}[t]
	\centering
	\includegraphics[width=0.90 \textwidth]{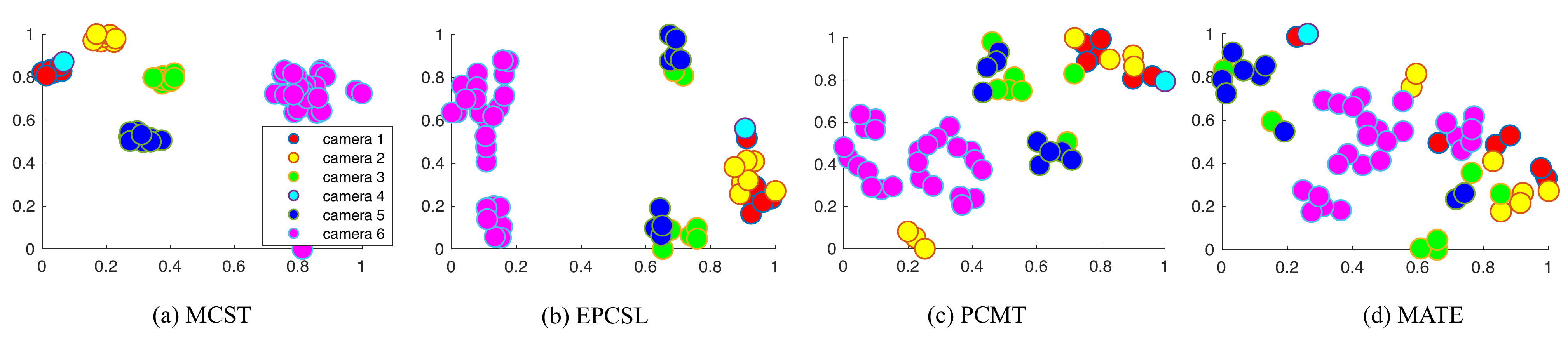}
	\caption{Feature distribution visualisation of a randomly selected person identity
		appearing under all the six camera views of the Market-1501 dataset.
		This is made by t-SNE \citep{maaten2008visualizing}.
		Camera views are colour-coded. Best viewed in colour.
	}
	\label{fig:three_baselines_tsne}
\end{figure*}

\subsection{Benchmarking the ICS Person Re-ID}
\label{subsec:baselines_ICS_reid}

Since there is no dedicated methods for solving the 
proposed ICS person re-id problem,
we formulated and benchmarked three baseline methods based on the 
generic learning algorithms:
\begin{enumerate}
\item {\em Multi-Camera Single-Task} (MCST) learning (Fig. \ref{fig:three_baselines}(a)):
Given no identity association across camera views,
we simply consider that identity classes from different camera views are 
distinct people and merge all the per-camera label spaces into a joint space cumulatively.
This enables the conventional supervised model learning based on identity classification.
We therefore train a single re-id model, 
as in the common supervised learning paradigm.
At test time, we extract the re-id feature vectors
and apply the Euclidean distance as the metrics for re-id matching.

\item {\em Ensemble of Per-Camera Supervised} (EPCS) learning (Fig. \ref{fig:three_baselines}(b)):
Without inter-camera identity labels, for each camera view we train a separate re-id model with its own single-camera training data. 
During deployment, given a test image we extract the feature vectors 
of all the per-camera models, concatenate them into a single representation vector, 
and utilise the Euclidean distance as the matching metrics for re-id.

\item {\em Per-Camera Multi-Task} (PCMT) learning (Fig. \ref{fig:three_baselines}(c)):
While being a variant of our \name{} model {\em without} the cross-camera multi-label learning component,
we simultaneously consider it as a baseline due to the use of the multi-task learning strategy.

\end{enumerate}

To implement fairly the baseline learning methods, we used the same backbone ResNet50 as our method,
a widely used architecture in the re-id literature. 
We trained each of these models with the softmax cross-entropy loss function
in their respective designs.

\vspace{0.1cm}
\noindent {\bf Results.}
We compared our \name{} model with the three baseline methods
in Table \ref{tab:benchmark}.
Several observations can be pointed: 
\begin{enumerate} 
\item Concatenating simply the per-camera identity label spaces,
MCST yields the weakest re-id performance.
This is not surprised because there is a large (unknown) proportion 
of duplicated identities but mistakenly labelled with different classes,
misleading the model training process. 
\item 
The above problem can be addressed by independently exploiting camera-specific identity class annotations, as EPCS does.
This method does produce better re-id model generalisation consistently. 
However, the over accuracy is still rather low, 
due to the incapability of leveraging the shared knowledge between camera views and mining the inter-camera identity matching information. 
\item 
To address this cross-camera association issue,
PCMT provides an implicit solution and significantly
improves the model performance.
\item Moreover, the proposed \name{} model further 
boosts the re-id matching accuracy by explicitly associating
the identity classes across camera views in a reliable formulation. 
This verifies the efficacy of our model in capitalising
such cheaper and more scalable per-camera identity labelling.
\end{enumerate}

To further examine the model performance,
in Fig. \ref{fig:three_baselines_tsne} we visualised the feature distributions 
of a randomly selected person identity
with images captured by all the camera views
of Market-1501.
It is shown that 
the feature points of our model present
the best camera-invariance property, qualitatively validating
the superior re-id performance over other competitors.

\begin{table*} 
	\center
	\setlength{\tabcolsep}{0.15cm}
	\caption{
		Comparative evaluation of representative person re-id paradigms
		in the model training {\em supervision} perspective.
		`$^\dagger$': Results from \citep{yu2019mar}.
		`$^*$': Results from \citep{xin2019semi}.
	}
	\label{tb:exp_comparsion_sota}
	\begin{tabular}{c||l||ccc|c||ccc|c||ccc|c}
		\hline
		\multirow{2}{*}{\em Supervision} & \multirow{2}{*}{Method}                       
		& \multicolumn{4}{C{3.2cm}||}{Market-1501} & \multicolumn{4}{C{3.2cm}||}{DukeMTMC-reID} & \multicolumn{4}{|C{3.2cm}}{MSMT17}\\
		\cline{3-14}
		&
		& R1 & R10 & R20 & mAP
		& R1 & R10 & R20 & mAP
		& R1 & R10 & R20 & mAP \\
		\hline \hline
		\multirow{4}{*}{None} 
		& RKSL$^\dagger$ 
		& 34.0 & - & - & 11.0
		& - & - & - & -
		& 15.4 & - & - & 4.3
		\\
		& ISR$^\dagger$ 
		& 40.3 & - & - & 14.3
		& - & - & - & -
		& 21.5 & - & - & 6.1 \\
		& DIC$^\dagger$ 
		& 50.2 & - & - & 22.7
		& - & - & - & - 
		& 22.8 & - & - & 7.0
		\\
		& BUC 
		& {66.2} & 84.5 & - & {38.3}  
		& {47.4} & 68.4 & - & {27.5} 
		& {-} & {-} & {-} & {-} 
		\\
		& TSSL 
		& 71.2 & - & - & 43.3
		& 62.2 & - & - & 38.5
		& {-} & {-} & {-} & {-}
		\\
		\hline \hline
		\multirow{2}{*}{Tracking} 
		& TAUDL 
		& 63.7 & - & - & 41.2
		& 61.7 & - & - & 43.5
		& - & - & - & - \\
		& UTAL 
		& {69.2} & 85.5 & 89.7 & {46.2}  
		& {62.3} & 80.7 & 84.4 & {44.6} 
		& {31.4} & 51.0 & 58.1  & {13.1} \\
		\hline \hline
		\multirow{5}{*}{Source Domain} 
		& CAMEL 
		&  {54.5} & - & - & {26.3} 
		& {-} & {-} & {-} & {-}
		& {-} & {-} & {-} & {-}\\
		& TJ-AIDL 
		& {58.2} & {-} & {-} & {26.5}
		& {44.3} & {-} & {-} & {23.0}
		& {-} & {-} & {-} & {-}\\
		& CR-GAN 
		& 59.6 & - & - & 29.6 
		& 52.2 & - & -  & 30.0
		& - & - & - & - \\
		& MAR 
		& 67.7 & - &- & 40.0
		& 67.1 & - & - & 48.0
		& - & - & - & - \\
		& ECN 
		& 75.1 & 91.6 & - & 43.0 
		& 63.3 & 80.4 & - & 40.4 
		& 30.2 & 46.8 & - & 10.2\\
		\hline \hline 
		Intra-Camera
		& \bf \name{} (Ours) 
		& {88.7} & {97.1} & {98.2} & {71.1}
		&  {76.9} & {89.6} & {92.3} & {56.6}
		&  {46.0} & {65.3} & {71.1} & {19.1}\\
		\hline
		\hline
		\multirow{3}{*}{Cross-Camera}
		& ResNet50$^*$
		& 66.1 & - & - & 42.1
		& 50.0 & - & - & 30.3
		& - & - & - & -
		\\
		& WRN50$^*$
		& 65.8 & - & - & 42.2 
		& 49.4 & - & - & 30.9
		& - & - & - & -
		\\ 
		(Semi) 
		& MVC
		& 72.2 & - & - & 49.6 
		& 52.9 & - & - & 33.6
		& - & - & - & -
		\\
		\hline
		\hline
		\multirow{5}{*}{Cross-Camera}
		& HA-CNN 
		& 91.2 & - & - & 75.7
		& 80.5 & - & - &  63.8 
		& - & - & - & - \\
		& SGGNN 
		& 92.3 & - & - & 82.8 
		& 81.1 & - & - & 68.2 
		& - & - & - & - \\
		& PCB
		& 93.8 & - & - & 81.6 
		& 83.3 & - & - & 69.2 
		& 68.2 & - & - & 40.4 \\
		& JDGL 
		& 94.8 & - & - & 86.0
		& 86.6 & - & - & 74.8
		& 77.2 & - & - & 52.3 \\
		& OSNet 
		& 94.8 & - & - & 84.9 
		& 88.6 & - & - & 73.5
		& 78.7 & - & - & 52.9 \\
		\hline 
	\end{tabular}
\end{table*}

\subsection{Comparing Different Person Re-ID Paradigms}

As a novel re-id person scenario, 
it is informative and necessary to compare 
with other existing scenarios in the problem-solving and 
supervision cost perspectives.
To this end, we compared ICS with existing representative re-id paradigms
in an increasing order of training supervision cost:
\begin{enumerate}
\item {\em Unsupervised learning} (no supervision):
RKSL \citep{wang2016towards},
ISR \citep{lisanti2014person},
DIC \citep{kodirov2015dictionary},
BUC \citep{lin2019BUC},
and 
TSSL \citep{wu2020TSSL};

\item {\em Tracking data modelling}:
TAUDL \citep{li2018taudl} and
UTAL \citep{li2019utal_pami};

\item {\em Unsupervised domain adaptation} (source domain supervision):
CAMEL \citep{uda_yu2017cross},
TJ-AIDL \citep{uda_wang2018transferable}, 
CR-GAN \citep{chen2019Instance},
MAR \citep{yu2019mar},
and ECN \citep{zhong2019ECN};

\item {\em Semi-supervised learning} (cross-camera supervision at small size):
ResNet50 \citep{net_arc2016resnet},
WRN50 \citep{zagoruyko2016wide}, and
MVC \citep{xin2019semi};

\item {\em Conventional fully supervised learning} (cross-camera supervision):
HA-CNN \citep{li2018harmonious},
SGGNN \citep{shen2018person},
PCB \citep{sun2018beyond},
JDGL \citep{zheng2019joint}, and
OSNet \citep{zhou2019osnet}.
\end{enumerate}

Table \ref{tb:exp_comparsion_sota} presents 
a comprehensive comparative evaluation of different person re-id paradigms 
in terms of the model performance {\em VS.} 
supervision requirement. We highlight the following 
observations:
\begin{enumerate}
\item 
Early unsupervised learning re-id models (RKSL, ISR, DIC),
which rely on hand-crafted visual feature representations,
often yield very limited re-id matching accuracy.
While deep learning clearly improves the performance as shown in BUC and TSSL,
the results are still largely unsatisfactory.

\item 
By exploiting tracking information including spatio-temporal object appearance continuity,
TAUDL and UTAL further improve the model generalisation.

\item 
Unsupervised domain adaptation is another classical approach to
eliminating the tedious collection of labelled training data per domain.
The key idea is knowledge transfer from a source dataset (domain)
with cross-camera labelled training samples.
This strategy continuously pushes up the matching accuracy.
It has the clear limitation that a {\em relevant} labelled source domain 
is assumed which however is not always guaranteed in practice. 

\item 
While semi-supervised learning enables label reduction,
the model performance remains unsatisfactory and 
is relatively inferior to unsupervised domain adaptation.
This paradigm relies on expensive cross-camera identity annotation
despite at smaller sizes.

\item 
With full cross-camera identity label supervision,
supervised learning methods produce the best re-id performance
among all the paradigms.
However, the need for cross-camera identity association
leads to very high labelling cost per domain, restricting significantly
its scalability in realistic large scale applications
typically with limited annotation budgets.
\item 
The ICS re-id is proposed exactly for solving this low cost-effectiveness limitation
of the conventional supervised learning re-id paradigm,
without the expensive cross-camera identity association labelling.
Despite much weaker supervision, \name{} can approach the performance
of the latest supervised learning re-id methods on Market-1501.
However, the performance gap on the largest dataset MSMT17 is still clearly bigger,
suggesting a large room for further ICS re-id algorithm innovations.
\end{enumerate}

\subsection{Further Evaluation of Our Method}
\label{subsec:eval_mtml}

We conducted a sequence of in-depth component evaluations
for the \name{} model 
on the Market-1501 dataset. 

\begin{table}[b]
	\setlength{\tabcolsep}{0.18cm}
	\caption{
		Evaluating the model components of \name{}:
		Per-Camera Multi-Task (PCMT) learning,
		Cross-Camera Multi-Label (CCML) learning,
		and Curriculum Thresholding (CT).
		Dataset: Market-1501.
	}
	\label{tb:ablation_study}
	\begin{tabular}{l|c|c|c|c}
		\hline
		\em Component 
		& R1 & R10 & R20 & mAP\\
		\hline
		\hline
		PCMT 
		& {78.4} & 93.1 & 95.7  & {52.1}\\
		PCMT+CCML 
		& {85.3} & {96.2} & {97.6} & {65.2} \\
		PCMT+CCML+CT ({\bf full}) 
		&\bf{88.7} & \bf{97.1} & \bf{98.2} &\bf {71.1} \\
		\hline 
	\end{tabular}
\end{table}

\begin{figure}[t!]
	\centering
	\includegraphics[width=0.35\textwidth]{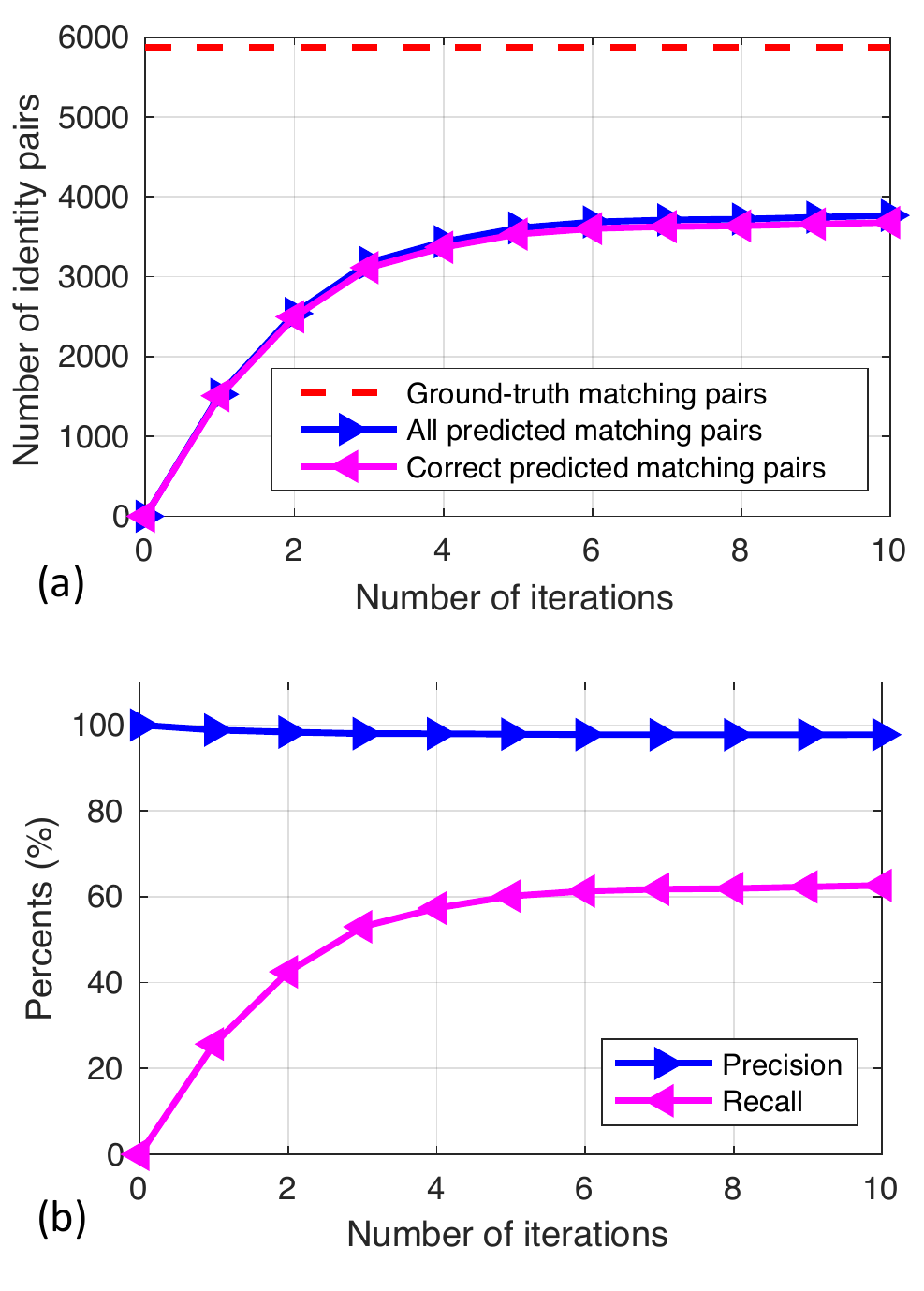}
	\caption{
		Dynamic statistics of cross-camera identity association 
		over the training rounds.
		Dataset: Market-1501.
		{\bf(a)} The number of ground-truth matching pairs ({\em ground-truth pairs}),
		the number of all predicted matching pairs ({\em all predicted pairs}), and
		the number of correctly predicted matching pairs ({\em correct predicted pairs}).
		{\bf(b)} The {\em precision} and {\em recall} of all predicted matching pairs.
	}
	\label{fig:ass_th05_lambda05}
\end{figure}

\begin{figure*}
	\centering
	\includegraphics[width=0.99\textwidth]{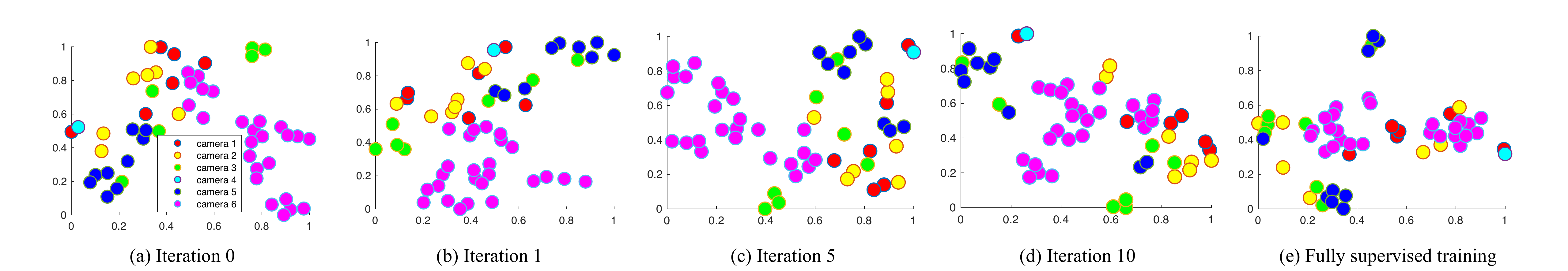}
	\caption{
		{\bf (a-d)} The feature distribution evolution of a set of multi-camera images from a single random person
		over the training rounds, in comparison to
		{\bf (e)} the feature distribution by supervised learning. 
		Iteration 0 indicates the initial feature space before training starts.
		Dataset: Market-1501. 
		Best viewed in colour.
	}
	\label{fig:tsne_six_cams_training_dynamics}
\end{figure*}

\begin{figure*} [t]
    \centering
    \includegraphics[width=0.99\textwidth]{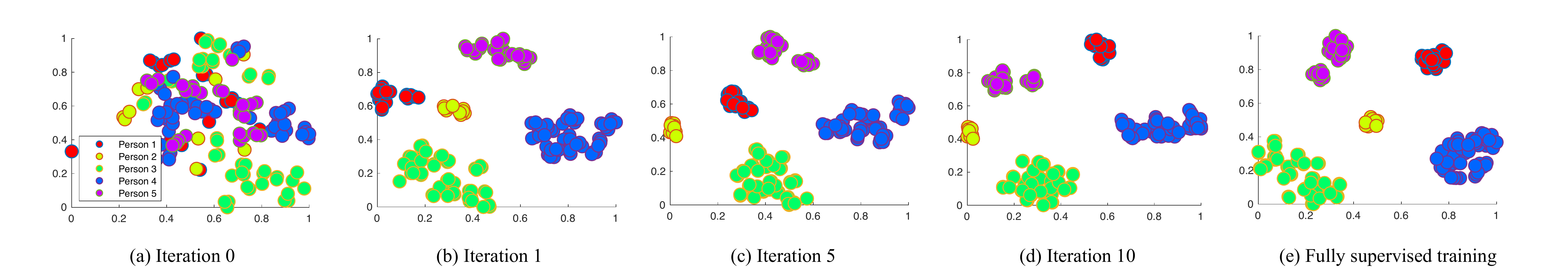}
    \caption{
    {\bf (a-d)} The feature distribution evolution of multi-camera images from five random persons 
		over the training rounds, in comparison to
		{\bf (e)} the feature distribution by supervised learning.
		Iteration 0 indicates the initial feature space before training starts.
		Dataset: Market-1501. 
		Best viewed in colour.
		}
    \label{fig:tsn_five_persons_training_dynamics}
\end{figure*}

\subsubsection{Ablation Study}
\label{subsubsec:ablation_study}
We started by evaluating the three components of
our \name{} model:
{\em Per-Camera Multi-Task} (PCMT) learning,
{\em Cross-Camera Multi-Label} (CCML) learning,
and
{\em Curriculum Thresholding} (CT).
The results in Table \ref{tb:ablation_study} show
that:
(1) Using the PCMT component alone, 
the model can already achieve fairly strong re-id matching performance,
thanks to the ability of learning implicitly cross-camera feature representation
via a specially designed multi-task inference structure. 
(2) Adding the CCML component significantly boosts the accuracy,
verifying the capability of our cross-camera identity matching strategy
in discovering the underlying image pairs.
(3) With the help of CT, a further performance gain is realised,
validating the idea of exploiting curriculum learning 
and the design of our curriculum threshold.

\begin{figure}[b]
    \centering
    \includegraphics[width=1.0\linewidth]{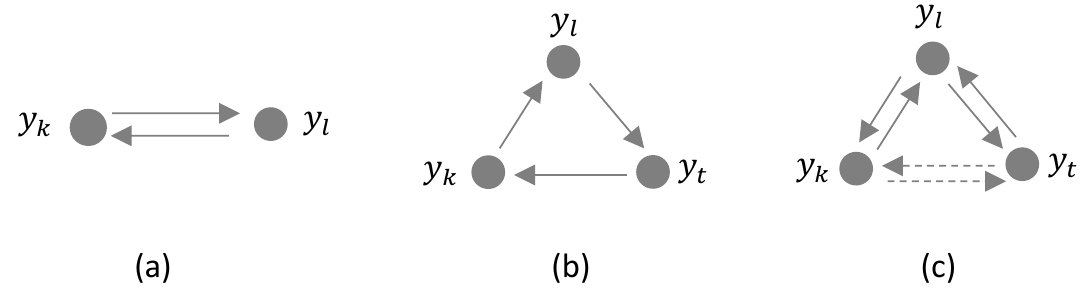}
    \caption{\revision{
    Illustration of three methods for identity cyclic association across camera views: 
    (a) Association across two camera views, adopted by \name; 
    (b) Association across three camera views; 
    (c) Transitive association across three camera views. 
    $y_k$, $y_l$, $y_t$ are three identities from different camera views. 
    Sold arrow denotes the correspondence relation discovered as in Sec. \ref{subsec:cross_cam_learning}, and dashed arrow in (c) denotes transitive association. 
    Both (b) and (c) can be further extended to more camera views.
    }}
    \label{fig:association_differences}
\end{figure}

\begin{figure}[t!]
    \centering
    \includegraphics[width=0.35\textwidth]{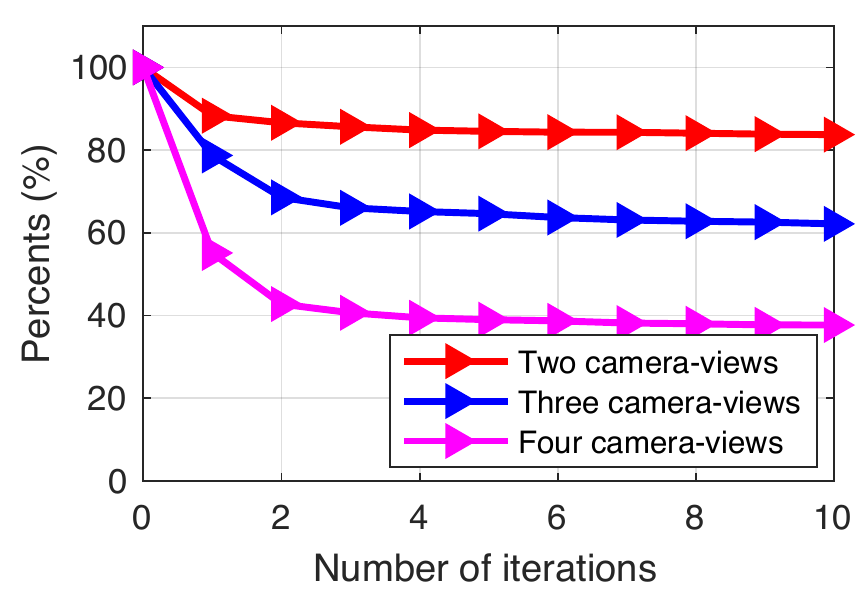}
    \caption{Comparing the association precision when varying numbers of camera views are involved in cyclic consistent association.}
    \label{fig:more_cam}
\end{figure}

As a key performance contributor, we further examined CCML by 
evaluating its essential part -- cross-camera identity association.
To this end, we tracked the statistics of self-discovered identity pairs across camera views over the training rounds, including the precision and recall measurements.
It is shown in Fig. \ref{fig:ass_th05_lambda05} that 
our model can mine an increasing number of identity association pairs
whilst maintaining very high precision
which therefore well limits the risk of error propagation and its disaster consequence. 
This explains the efficacy of our cross-camera multi-label learning.
On the other hand, 
while failing to identify around 40\% identity pairs,
our model can still achieve very competitive performance
as compared to fully supervised learning models.
This suggests that our method has already discovered the majority 
of re-id discrimination information from the associated identity pairs,
missing only a small fraction embedded in those hard-to-match pairs.
In this regard, we consider the proposed model is making a satisfactory trade-off between identity association error
and knowledge mining.  
\revision{
To check the impact of cross-camera identity association together with
per-camera learning, we visualised the change of feature distribution during training.
For a set of multi-camera images from a single person,
it is observed in Fig. \ref{fig:tsne_six_cams_training_dynamics}
that they are associated gradually
in the re-id feature space, reaching a similar distribution
as in the supervised learning case. 
For a set of images from five random persons,
our model enables them to be gradually pushed away,
as shown in Fig. \ref{fig:tsn_five_persons_training_dynamics}.
}
These observations are in line with the numerical performance evaluation
above.

\begin{figure*}[t!]
	\centering
	\includegraphics[width=0.95\textwidth]{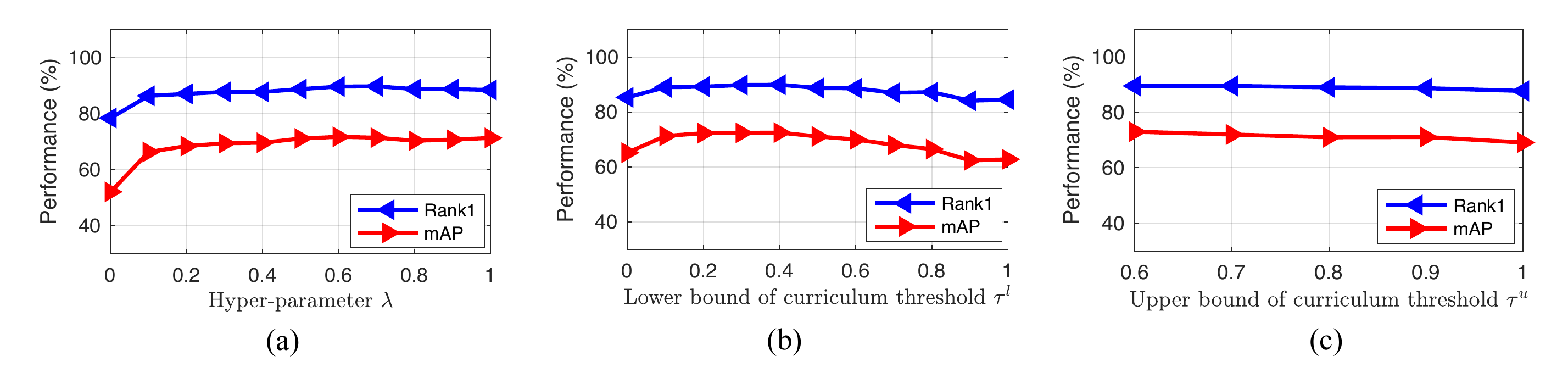}
	\caption{
		Hyper-parameter analysis:
		(\textbf{a}) the loss weight $\lambda$ in Eq. \eqref{eq:mmm_loss}, 
		the (\textbf{b}) lower and (\textbf{c}) upper bound of curriculum threshold in Eq. \eqref{eq:threshold}.
		Dataset: Market-1501.
	}
	\label{fig:sens_analysis}
\end{figure*}

\vspace{0.2cm}
\noindent \revision{{\bf Associative scope. }
Conceptually, the proposed concept of
cyclic consistent association can be extended to three or more camera views.
An example for three camera views is illustrated 
in Fig. \ref{fig:association_differences}(b).
For a more focused evaluation, we analysed this aspect
without curriculum threshold.
We considered 2, 3, and 4 camera views involved 
during association.
We obtained Rank-1/mAP rates of
85.3\%/65.2\%, 
{83.5\%/64.2\%},  
and {80.7\%/58.9\%},
respectively.
This result shows that the more camera views involved,
the lower model performance obtained.
The plausible reason is
that the negative effect of error propagation would be amplified
when additional camera views are added
into the associating cycle.
This is clearly reflected 
in the comparison of association precision, as shown in Fig. \ref{fig:more_cam}.
}

\vspace{0.2cm}
\noindent \revision{{\bf Transitive association.}
As shown in Fig. \ref{fig:association_differences}(c), transitive association means that if two identities ($y_k$ and $y_t$) are both associated with another identity ($y_l$)
in a cross-camera sense, 
then the two identities $y_k$ and $y_t$ should be also associated.
In \name{}, the transitive association is implicitly considered.
More specifically, when $y_k$ and $y_t$ both are pulled close towards $y_l$ concurrently, $y_k$ and $y_t$ will be made close in feature space during training,
i.e. $y_k$ and $y_t$ are associated.
This transitive association can be further extended to 4 or more camera views.
To verify the above analysis, 
we evaluated the effect of explicitly exploiting the transitivity information in training MATE.
We obtained 88.9\%/71.2\% in R1/mAP, 
similar to the performance of 88.7\%/71.1\% when it is implicitly utilised.
In design, we finally choose to implicitly mine such transitive relations for reduced model complexity.
}

\subsubsection{Hyper-Parameter Analysis}
\label{subsubsec:para_analysis}
We examined the performance sensitivity of three parameters of \name{}:
the loss weight $\lambda$ (default value 0.5) in Eq. \eqref{eq:mmm_loss},
the lower (default value 0.5) and upper (default value 0.95) bound of curriculum threshold in Eq. \eqref{eq:threshold}.
\revision{We evaluated each individual parameter by varying its value while setting all the others to their default values.}
Fig. \ref{fig:sens_analysis}
shows that all these parameters have a wide range of 
satisfactory values in terms of performance.
This suggests the ease and convenience of setting up model training
and good accuracy stability of our method.

\subsection{Intra-Camera Annotation Cost}

\revision{
We conducted a controlled data annotation experiment to annotate intra-camera person identity labels on the MSMT17 dataset \citep{dataset2018msmt17}.
Specifically, we annotated the identity labels of person images in a camera-independent manner,
with the original identity information discarded.
Due to the nature of per-camera person labelling, 
the entire identity space is split into
multiple independent, smaller spaces.
This allows us to decompose the labelling task
easily and enable multiple annotators
to conduct the labelling job in parallel without 
any interference and conflict among them.
These merits reduce significantly the annotation cost.

{\revisionTwo 
We { provide} a quantitative comparison on {the} annotation
{costs between}	
ICS and {a conventional fully supervised person re-id setting}. 
This {experiment} was {performed} on a subset of MSMT17.
%
Specifically, we randomly selected up to $50$ persons from each camera-view
which gives rise to a total of 714 identities.
We asked {three annotators to label the images}
using the same labelling tool we developed.
In annotating fully supervised data, the annotation is performed without using camera information. 
The labelling costs of ICS and the fully supervised {setting
are respectively} 2.5 and 8 person-days.
This {empirical validation is largely consistent with our annotation
{cost} complexity analysis {provided in the Introduction section}.
This experiment {demonstrates} that our ICS setting is
more efficient and more scalable {by reducing the annotation
  complexity and costs}.}}

In terms of performance, our method achieves 
a Rank-1/mAP rate of 46.0\%/19.1\%,
{\em vs.} 78.7\%/52.9\% by
the best supervised learning model
OSNet, whilst clearly outperforming all unsupervised, tracking, domain adaptation based alternatives (cf. Table \ref{tab:benchmark}).
This is an encouraging preliminary effort
of intra-camera supervised person re-id,
with a good improvement space remaining
in algorithm innovation.
}

\section{Conclusions}
In this work, we presented 
a novel person re-identification paradigm, i.e. intra-camera supervised (ICS) learning,
characterised by training re-id models
with only per-camera independent person
identity labels, 
but no the conventional cross-camera identity labelling. 
%
The key motivation lies in eliminating the tedious and expensive process of
manually associating identity classes across every pair of
camera views in a surveillance network, which makes the training data collection
too costly to be affordable in large real-world application. 
To address the ICS re-id problem, we formulated a \nameFull{}
(\name{}) learning model capable of fully exploiting per-camera re-id supervision 
whilst simultaneously self-discovering cross-camera
identity association.
Extensive evaluations were conducted on three re-id benchmarks to validate the advantages
of the proposed \name{} model over the state-of-the-art alternative methods in the proposed ICS learning setting.
Detailed ablation analysis is also provided for giving insights on our model design.
We conducted extensive comparative evaluations
to demonstrate the cost-effectiveness advantages of the ICS re-id paradigm 
over a wide range of existing representative re-id settings
and the performance superiority of our \name{} model
over alternative learning methods.

\section*{Acknowledgement}
This work was partially supported by Vision Semantics Limited, the Alan 
	Turing Institute Fellowship Project on Deep Learning for Large-Scale 
	Video Semantic Search, and the Innovate UK Industrial Challenge Project 
	on Developing and Commercialising Intelligent Video Analytics Solutions 
	for Public Safety (98111-571149).

\bibliographystyle{spbasic}      
\bibliography{ws_reid}   

\end{document}